\documentclass[10pt,twocolumn,letterpaper]{article}

\usepackage[pagenumbers]{cvpr} 

\PassOptionsToPackage{table,dvipsnames}{xcolor}
\usepackage{caption}
\usepackage{subcaption}
\usepackage{outlines}
\usepackage{cuted}
\usepackage{xcolor}
\usepackage{colortbl}
\usepackage{bm}
\usepackage{array}
\usepackage{algpseudocode}
\usepackage{algorithm}
\usepackage{booktabs}
\usepackage{multirow}
\usepackage{amssymb}
\usepackage{tabularx}

\definecolor{cvprblue}{rgb}{0.21,0.49,0.74}
\usepackage[pagebackref,breaklinks,colorlinks,allcolors=cvprblue]{hyperref}

\usepackage[capitalize]{cleveref}
\crefname{section}{Sec.}{Secs.}
\Crefname{section}{Section}{Sections}
\Crefname{table}{Table}{Tables}
\crefname{table}{Tab.}{Tabs.}
\definecolor{Cerulean}{rgb}{0.0, 0.48, 0.65}
\definecolor{myred}{rgb}{1, 0.6, 0.6}
\definecolor{myyellow}{rgb}{1,1, 0.6}
\definecolor{myorange}{rgb}{1, 0.8, 0.6}
\definecolor{mycolor_blue}{HTML}{E7EFFA}
\definecolor{mycolor_green}{HTML}{E6F8E0}
\definecolor{mycolor_gray}{HTML}{ECECEC}
\definecolor{pearDark}{HTML}{2980B9}
\newcommand{\tablefirst}[0]{\cellcolor{pearDark!20}}
\newcommand{\tablesecond}[0]{\cellcolor{mycolor_green}}

\def\paperID{1633} 
\def\confName{CVPR}
\def\confYear{2025}

\title{MEAT: Multiview Diffusion Model for \\ Human Generation on Megapixels with \underline{Me}sh \underline{At}tention}

\author{Yuhan Wang$^{1}$\quad 
Fangzhou Hong$^{1}$\quad 
Shuai Yang$^{2}$\quad 
Liming Jiang$^{1}$ \quad 
Wayne Wu$^{3}$ \quad
Chen Change Loy$^{1}$ \\
$^1$S-Lab, Nanyang Technological University \qquad $^2$WICT, Peking University \qquad $^3$UCLA \\
}

\begin{document}

\twocolumn[{
    \renewcommand\twocolumn[1][]{#1}
    \vspace{-1em}
    \maketitle
    \vspace{-1em}
    \begin{center}
        \vspace{-20pt}
        \centering
        \includegraphics[width=0.98\textwidth]{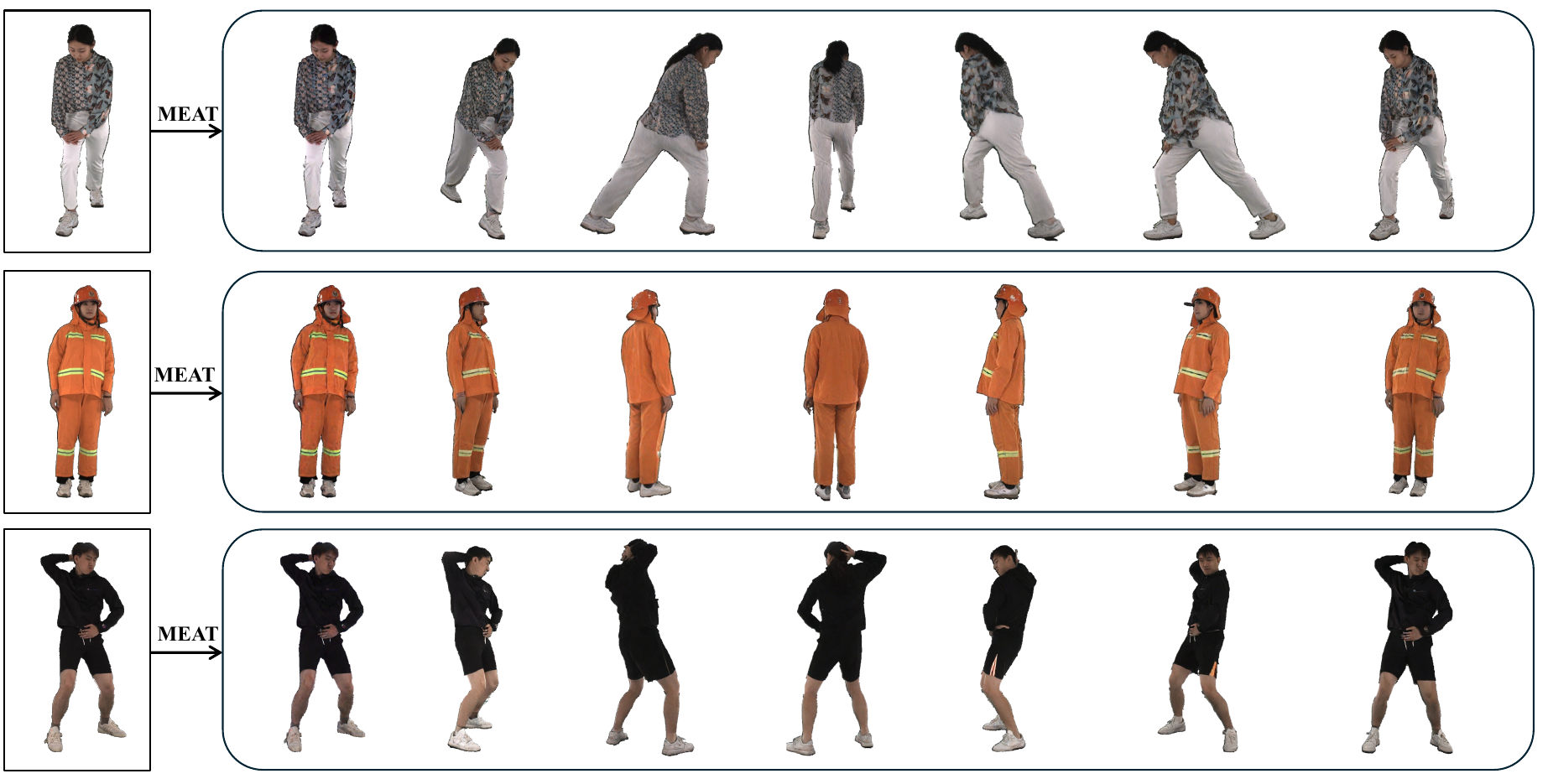}
        \vspace{-9pt}
        \captionof{figure} {
        Given a frontal human image, \textbf{MEAT} can generate dense, view-consistent multiview images at a resolution of $1024^2$.
        }
        \label{fig:teaser}
    \end{center}
}]


\begin{abstract}
Multiview diffusion models have shown considerable success in image-to-3D generation for general objects. However, when applied to human data, existing methods have yet to deliver promising results, largely due to the challenges of scaling multiview attention to higher resolutions.
In this paper, we explore human multiview diffusion models at the megapixel level and introduce a solution called \textbf{mesh attention} to enable training at 1024$^2$ resolution.
Using a clothed human mesh as a central coarse geometric representation, the proposed mesh attention leverages rasterization and projection to establish direct cross-view coordinate correspondences. This approach significantly reduces the complexity of multiview attention while maintaining cross-view consistency.
Building on this foundation, we devise a mesh attention block and combine it with keypoint conditioning to create our human-specific multiview diffusion model, \textbf{MEAT}.
In addition, we present valuable insights into applying multiview human motion videos for diffusion training, addressing the longstanding issue of data scarcity.
Extensive experiments show that MEAT effectively generates dense, consistent multiview human images at the megapixel level, outperforming existing multiview diffusion methods. Code is available at \href{https://johann.wang/MEAT/}{https://johann.wang/MEAT/}.
\end{abstract}    
\vspace{-25pt}
\section{Introduction}
\label{sec:intro}

\begin{figure}[t]
    \centering
    \includegraphics[width=\columnwidth]{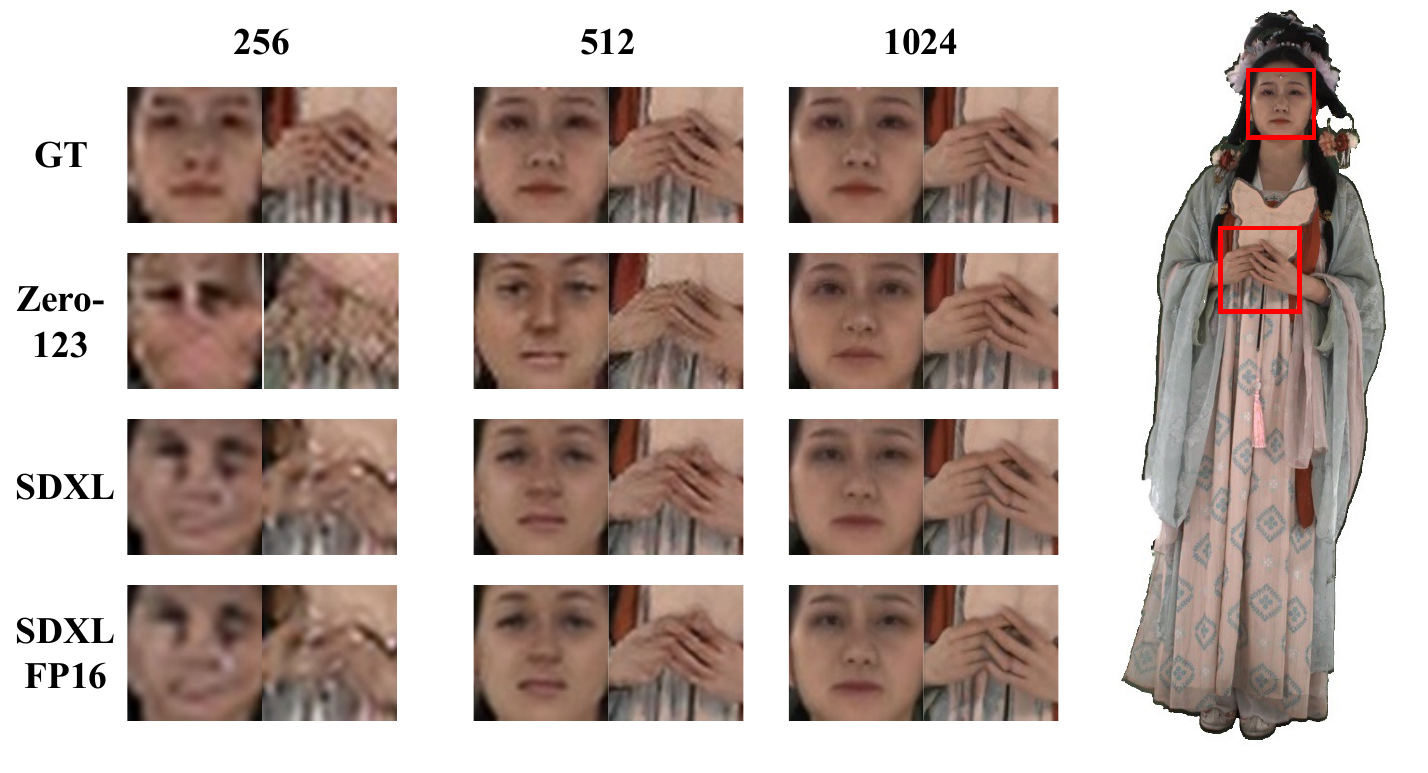}
    \caption{\textbf{VAE and Resolution.} Each row represents the same version of VAE, while each column corresponds to the same resolution of the full-body image after VAE reconstruction. 
    Although the full-body image rendered at $512\times512$ shows good visual quality, it falls short when used in diffusion models with VAE. We find that a resolution of $1024\times1024$ is necessary for optimal results.
    }
    \label{fig:vae_res}
\end{figure}

In this paper, we address the problem of multiview human generation, which aims to generate realistic, consistent multi-angle renderings of a human figure. We assume a single frontal image is provided.
Recent advancements in diffusion models offer a promising new approach to this task, as they excel at generating high-quality images conditioned on various inputs. 
However, achieving realistic human renderings remains highly challenging due to the importance of resolution for capturing fine details. Specifically, existing multiview diffusion models \cite{zero1to3, shi2023zero123++, long2024wonder3d, huang2024epidiff} for general objects are typically trained at a resolution of $256^2$, with a few recent methods increasing this to $512^2$ \cite{li2024era3d} or $578^2$ \cite{voleti2025sv3d}. However, this remains insufficient for human data. As shown in Fig.~\ref{fig:vae_res}, under the latent diffusion setting, a resolution of $1024^2$ is necessary to achieve satisfactory results, as the result is highly sensitive to details in areas such as the face, hands, and clothing. Any lack of detail, unnatural appearance, or inconsistency in these regions significantly diminishes the realism. Since these areas each occupy only a small portion of the overall pixel space, variational autoencoder (VAE) reconstructions at resolutions below $1024\times1024$ are suboptimal, making it challenging to train an effective multiview diffusion model.

\begingroup
\setlength{\tabcolsep}{4.4pt}
\begin{table}[t]
\caption{\textbf{Multiview Attention Comparison.} (1) Dense multiview attention requires each pixel to integrate all other pixels in different views, consuming $N\times$ more memory than self-attention. (2) Row-wise attention is based on the orthographic assumption, making it unsuitable for videos shot with an arbitrary perspective. (3) Epipolar attention is related to our approach. It requires sampling 3D point candidates for each pixel, with the density $K$ balancing multiview accuracy and complexity. (4) Our mesh attention eliminates this sampling with a centric mesh. We assume the feature map dimensions are  $H=W=S$, with each view interacting with all $N$ views. $d$ represents the grid sampling constant.
}
\centering
{\fontsize{8pt}{11pt}\selectfont
\begin{tabularx}{\columnwidth}{l|ccc|c}
\toprule
\textbf{Attn. Type}  & \textbf{Q}       & \textbf{K,V}         & \textbf{Attn. Map}                              & \textbf{Persp.} \\ \midrule
Self-Attn  & $NCS^2$  & $NCS^2$      & $NS^4$                      & -               \\ \midrule
Dense MV & $NCS^2$  & $NC(NS^2)$     & $N^2S^4$   & \checkmark             \\
Row-wise        & $(NH)CW$  & $(NH)C(NW)$   & $N^2S^3$   & $\times$              \\
Epipolar        & $(NS^2)C\cdot1$ & $(NS^2)C(NKd)$ & $N^2S^2Kd$ & \checkmark             \\
\textbf{Mesh Attn}  & $(NS^2)C\cdot1$ & $(NS^2)C(Nd)$  & $\bm{N^2S^2d}$  & \checkmark \\ \bottomrule
\end{tabularx}
}
\label{tab:multiview_attn}
\end{table}
\endgroup

Directly increasing the working resolution of existing multiview diffusion models to $1024\times1024$ is impractical either. To maintain multiview consistency, current methods generate all views simultaneously and add cross-view attention within the denoising U-Net to integrate features from different views.
\Cref{tab:multiview_attn} summarizes the attention map complexity of existing multiview attention methods. Dense multiview attention \cite{shimvdream, wang2023imagedream, long2024wonder3d} has extremely high memory requirements, making it difficult to apply directly at megapixel resolutions. Meanwhile, row-wise attention \cite{li2024era3d} relies on an orthographic projection assumption, which significantly restricts the applicable training data.

To address these challenges, we propose MEAT, a multiview diffusion model designed for human novel view generation on megapixels, conditioned on a frontal image. 
In particular, we wish to address the high computational complexity of multiview attention in existing diffusion models. 
Our key idea is to leverage a rough central 3D representation that enables our method to directly establish correspondences between pixels across different viewpoints using rasterization and projection. We refer to this pixel-correspondence-based feature fusion as \textbf{mesh attention}. It allows us to sample sufficiently dense viewpoints on each GPU and train the model using $1024\times1024$ images.
As shown in Table \ref{tab:multiview_attn}, our method achieves the lowest complexity and offers graceful complexity growth as resolution increases.
Building on the design principles of Zero-1-to-3~\cite{zero1to3}, we generate all target views in parallel and introduce mesh attention blocks to maintain cross-view consistency. In addition, we enhance texture and geometric consistency by incorporating multi-scale VAE latent features and keypoints conditioning.

Apart from introducing the MEAT approach, we also present a new training source. The typical data source for multiview diffusion models is textured mesh data. However, high-quality human scan data at $1024\times1024$ resolution is extremely scarce and mostly limited to static poses. Even the largest dataset, THUman2.1 \cite{tao2021function4d_thuman}, includes only around 2,500 multiview subjects, making multiview diffusion model training highly susceptible to overfitting. 
To address this, we propose a data processing pipeline that leverages DNA-Rendering \cite{cheng2023dnarendering}, a multiview human motion video dataset, as a training source. The data greatly increases the diversity of poses available during training. We will discuss a series of techniques for adapting this dataset to train our mesh-attention-based multiview diffusion model.

To summarize, our main contributions are as follows:
\begin{itemize}
    \item We propose mesh attention, which establishes correspondences between pixels using rasterization and projection of a centric mesh, making it the most efficient cross-view attention method to date.
    \item Based on mesh attention, we introduce a human-specific multiview diffusion model, MEAT, capable of generating consistent 16-view images at megapixel resolution.
    \item We present techniques for adapting a large-scale multi-view human motion video dataset as a training source for multiview diffusion.
\end{itemize}
\section{Related Work}
\label{sec:related_works}

\noindent\textbf{Multiview Diffusion.} Research of multiview diffusion models began with Zero-1-to-3 \cite{zero1to3}, which first proposed using camera viewpoints as control conditions for image diffusion models to achieve novel view synthesis. As a one-view-at-a-time approach, it often produces inconsistencies in the generated views due to the stochastic nature of diffusion models. Subsequent approaches shifted to all-view-at-once generation to mitigate the inconsistency issue.

The first category of methods \cite{tang2023mvdiffusion, shimvdream, wang2023imagedream, long2024wonder3d, li2024era3d, huang2024epidiff} treats the generation of each view as a separate branch of image generation, using multiview attention across branches to achieve feature fusion and consistency constraints. 
MVDream \cite{shimvdream} introduces dense multiview attention for single-object text-to-multiview generation. ImageDream \cite{wang2023imagedream} expands this approach to image-conditioned generation. Wonder3D \cite{long2024wonder3d} incorporates normal data and cross-domain attention to enhance geometric consistency. Recent methods have started optimizing the complexity of multiview attention. EpiDiff \cite{huang2024epidiff} uses epipolar attention for efficient pixel-matching candidate retrieval. Era3D \cite{li2024era3d} proposes row-wise attention based on the orthographic projection assumption. 
Other methods treat multiview images in alternative forms, such as a tiled big image \cite{shi2023zero123++} or a video \cite{voleti2025sv3d, gao2024cat3d}, leading to different approaches.
Our work, MEAT, further extends parallel multiview generation by enabling direct cross-view feature integration through rasterization and projection using a central 3D mesh representation.

\noindent\textbf{Monocular Human Reconstruction.} Monocular human reconstruction methods can be categorized into two groups based on whether they rely on optimizing a 3D representation. Optimization-based approaches, like ICON \cite{xiu2022icon} and ECON \cite{xiu2023econ}, achieve purely geometric clothed human reconstruction with aligned SMPL-X \cite{smplx} parameters, while TeCH \cite{huang2024tech} and SIFU \cite{zhang2024sifu} additionally support faithful texture generation. The other category of methods \cite{saito2019pifu, saito2020pifuhd, zheng2021pamir} use feed-forward networks to estimate the 3D occupancy field and extract the human mesh using the Marching Cubes algorithm \cite{lorensen1987marching_cube}, then attach textures through shape-guided inpainting \cite{albahar2023humansgd}.
A concurrent work, MagicMan, like our approach, combines a 512-resolution multiview diffusion model with monocular human reconstruction. MagicMan and our MEAT can generate dense multiview results that can be directly applied to 2DGS \cite{huang20242dgs} reconstruction.
\section{Methodology}
\label{sec:method}

\begin{figure*}[t]
    \vspace{-0.5cm}
    \centering
    \includegraphics[width=0.9\textwidth]{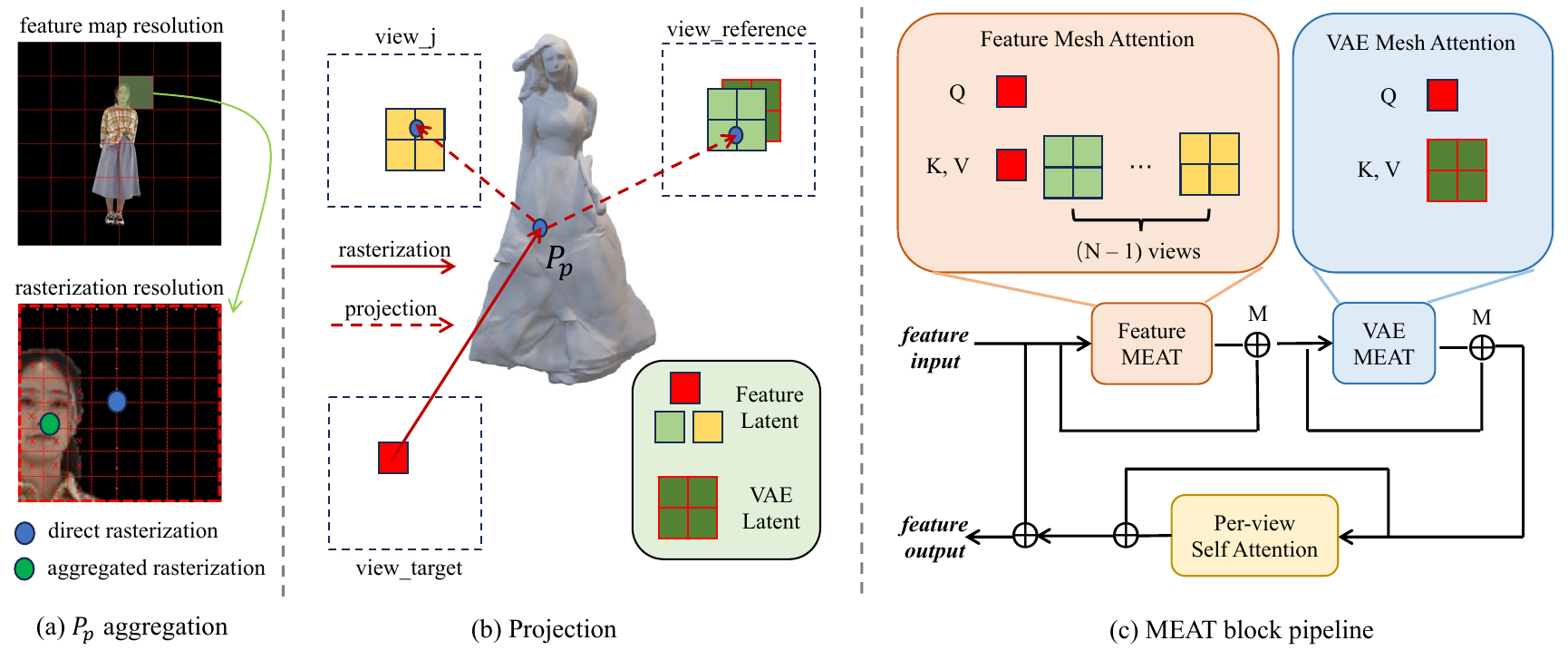}
    \caption{\textbf{Mesh Attention Block.} (a) $P_p$ aggregation. When the resolution of the feature map is very low, the ray cast from the center of a pixel may not intersect with the mesh, although the pixel area itself overlaps with it. (b) Projection. Each projected point is rounded to four integer pixels, corresponding to $d=4$ in \Cref{tab:multiview_attn}. The projected points on the reference view are also used to retrieve the encoded VAE features. (c) MEAT block pipeline. We use mesh attention to fuse U-Net features from all $N$ views, and VAE features from the reference. An additional per-view self-attention block is applied to process the captured multiview features. $M$ stands for masked skip connection.}
    \vspace{-0.2cm}
    \label{fig:meat}
\end{figure*}

\subsection{Preliminaries}
\label{subsec:method.preliminaries}

\noindent\textbf{Multiview Diffusion Models.} Existing multiview diffusion models typically consist of a VAE encoder $\mathcal{E}$, a denoiser U-Net $\epsilon_{\theta}$, and a VAE decoder $\mathcal{D}$. They can be categorized into two main types: one-view-at-a-time approaches~\cite{zero1to3, shi2023zero123++} and all-view-at-once methods~\cite{shimvdream, wang2023imagedream, huang2024epidiff, zheng2024free3d, long2024wonder3d, li2024era3d}.

The first category conditions the generation of \textbf{one} target view on the reference image $y$ and the relative camera rotation $R$ and translation $T$. The training objective is
\begin{equation}
    \label{eq:zero123}
    \min_{\theta}\mathbb{E}_{\mathcal{E}(x_{0}),\epsilon\sim\mathcal{N}(0,I),t} \Bigl[\lVert \epsilon-\epsilon_{\theta}(z_{t},t,y,R,T) \rVert_{2}^{2}\Bigr].
\end{equation}
Such models can generate multiview results sequentially but lack explicit consistency constraints across views.

The second category processes all views simultaneously:
\begin{equation}
    \label{eq:mvdiffusion}
    \min_{\theta}\mathbb{E}_{\mathcal{E}(x_{0}^{1:N}),\epsilon,t}\Bigl[\lVert \epsilon-\epsilon_{\theta}(z_{t}^{1:N},t,y,[R,T]^{1:N}) \rVert_{2}^{2}\Bigr].
\end{equation}
These methods yield better cross-view consistency at the cost of significant memory and computation overhead during the cross-view attention modules. As part of this type, our model efficiently produces nontrivial dense, $1024\times1024$ high-resolution multiview generation through a novel mesh attention mechanism, which we detail in \cref{subsec:method.mesh_attention}.

\noindent\textbf{Rasterization.} In mesh-based rasterization, each pixel on the 2D image plane is associated with a ray cast from the camera into 3D space, intersecting with the mesh surface. For each pixel $\bm{p}$, it computes the intersection mask $M_{\bm{p}}$, the intersected face index $\phi$, and the barycentric coordinates $\bm{\lambda_{p}}=(\lambda_{p1}, \lambda_{p2}, \lambda_{p3})$.
With the barycentric coordinates $\bm{\lambda_p}$ and the triangle face vertex coordinates $\bm{P}_{\phi}$, we can derive the 3D coordinates of the intersected point on mesh
\begin{equation}
    \label{eq:intersection}
    \bm{P_p} = \text{interp}(\bm{\lambda_p},\bm{P}_{\phi}).
\end{equation}
Our mesh attention takes advantage of the aggregation and projection of $\bm{P_p}$.

\subsection{Mesh Attention}
\label{subsec:method.mesh_attention}

We introduce mesh attention, MEAT, to overcome the inefficiencies of traditional cross-view attention, where each pixel must access and integrate information from all other pixels in different views, resulting in substantial redundant computation. In practice, pixels across views correspond to each other according to the 3D structure of the object.
Given an approximate clothed mesh as the centric coarse geometric representation of the human object, our approach leverages the 3D coordinate transformations to directly identify corresponding 2D pixel locations across different views. 
This allows us to aggregate information from these matched pixels, reducing redundancy and improving cross-view consistency. Details of MEAT are explained below.

\noindent \textbf{Aggregated Rasterization.} We can obtain the 3D coordinates of the intersection on the mesh for each pixel $\bm{p}$ through rasterization and Eq.~(\ref{eq:intersection}). However, due to the potentially low resolution of the diffusion features (\eg, $16\times16$ mid-block feature maps for $1024^2$ images), pixels near the object edges, which may contain useful information, can be misclassified as having no intersection with the mesh when using direct rasterization, as shown in Fig.~\ref{fig:meat}(a). To address this, we aggregate higher-resolution rasterization results to obtain the intersection point $\bm{P_p}$ and the mask $M_{\bm{p}}$ at the resolution of the feature map.

Consider a pixel $\bm{p}$ on the feature map that corresponds to a pixel region $S$ in the higher-resolution rasterization. We treat $\bm{P_p}$ as the average of all valid $\bm{P_s}$ within the region $S$:
\begin{align}
    \label{eq:meat_agg}
    \bm{P_p} &= \frac{\sum_{\bm{s}\in S}{M_{\bm{s}} \bm{P_s}}}{\sum_{\bm{s}\in S}M_{\bm{s}}}, \\
    M_{\bm{p}} &= \vee_{\bm{s}\in S}M_{\bm{s}},
\end{align}
where $\vee$ is the ``logical or'' operation.
The higher-resolution rasterization only needs to be performed once and can be reused for aggregation at different target resolutions.

\noindent \textbf{Projection and Grid Sampling.} After obtaining $\bm{P_p}$ for a target view pixel $\bm{p}$, we can use the calibration matrices $K_v, R_v, T_v$ of each view $v$ to locate the corresponding pixel of $\bm{P_p}$ in other views: 
\begin{equation}
    \label{eq:projection}
    \bm{p}_v = [K_v(R_v \bm{P_p} + T_v)]_{xy}.
\end{equation}
The corresponding features can then be retrieved using grid sampling. Instead of interpolating the features of neighboring pixels based on $\bm{p}_v=(x,y)$, we round $x, y$ up and down to extract the corresponding four features $\bm{f}_{v}$ from the feature map $\bm{F}_{v}$ of view $v$:
\begin{equation}
    \label{eq:grid_sample}
    \small
    \bm{f}_{v} = \mathrm{grid\_sample}(\bm{F}_{v},\{\lfloor x \rfloor, \lceil x \rceil\} \times \{\lfloor y \rfloor, \lceil y \rceil\}).
\end{equation}

\noindent \textbf{Cross-view Attention.} For pixel $\bm{p}$ on the target view with U-Net feature $\bm{f}$, we use cross attention to fuse the features from other views. To provide location priors, we concatenate the harmonic-embedded view camera pose $\bm{c}_v$ to the raw U-Net features $\bm{f}_{v}$. The masked skip connections are applied to omit pixels that do not intersect with the mesh from participating in mesh attention.
{\small
\begin{align}
    \label{eq:feat_meat}
    &Q=W_Q(\bm{f}\oplus \bm{c}_{tgt})\quad K,V=W_{K,V}(\bm{f}_{1:N}\oplus \bm{c}_{1:N}), \\
    &\mathrm{MEAT}_{feat}(\bm{f},\bm{p}) = M_{\bm{p}} \cdot \mathrm{Attention}(Q,K,T) + \bm{f},
\end{align}
}%
where $\oplus$ denotes channel-wise concatenation.

In addition to the fusion of U-Net features across views, we use a fully convolutional residual encoder to process VAE latent $z_0$ of the reference view into multi-scale feature tensors $\bm{F}_{\gamma}$ and inject them through mesh attention. Specifically, for pixel $\bm{p}$ on the target view, we use the projection of $\bm{P_p}$ on the reference view as the pixel location $\bm{p}_{ref}$ and employ grid sampling as defined in Eq.~(\ref{eq:grid_sample}) to extract $\bm{f}_{\gamma}$ from the VAE features. Mesh attention is applied exclusively to the reference view in this step.
{\small
\begin{align}
    \label{eq:vae_meat}
    &Q_{\gamma}=W_{Q_{\gamma}}(\bm{f}\oplus \bm{c}_{tgt})\quad K_{\gamma},V_{\gamma}=W_{K_{\gamma},V_{\gamma}}(\bm{f}_{\gamma}\oplus \bm{c}_{ref}), \\
    &\mathrm{MEAT}_{vae}(\bm{f},p) = M_p \cdot \mathrm{Attention}(Q_{\gamma},K_{\gamma},V_{\gamma}) + \bm{f}.
\end{align}
}%
Here, $ref$ and $tgt$ indicate the reference and target view. The above operations are applied to each pixel of each view.

After the two per-pixel attention operations, we apply a self-attention mechanism for each view to process the fused features. The complete pipeline is shown in Fig.~\ref{fig:meat}(c). 
In the classifier-free guidance training scheme, we always retain the mesh attention module and set 15\% of the data's camera embeddings and concatenated $x_0$ to null,  encouraging the model to fully leverage the mesh attention. 

\subsection{Multiview Diffusion Model with MEAT}
\label{subsec:method.human_multiview_diffusion}

\begin{figure}[t]
    \vspace{-0.4cm}
    \centering
    \includegraphics[width=0.95\columnwidth]{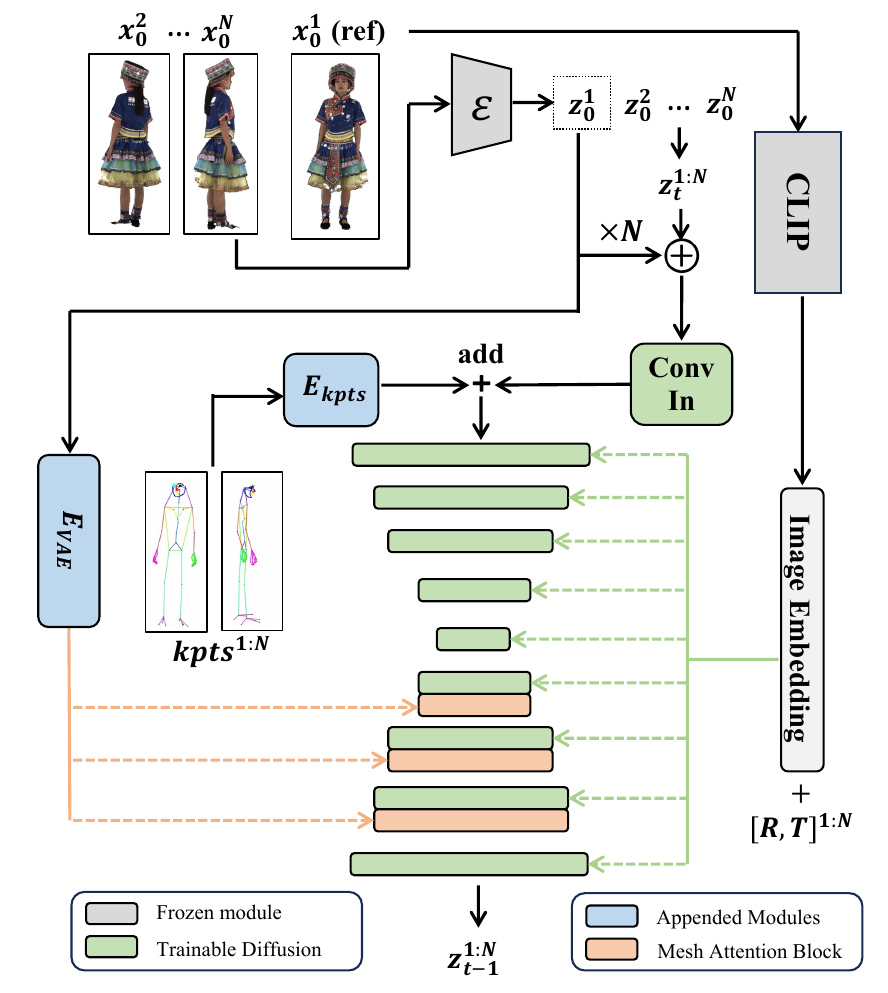}
    \caption{\textbf{Pipeline of MEAT.} We insert mesh attention blocks into up-sampling blocks of the U-Net to fuse multiview features.}
    \vspace{-0.4cm}
    \label{fig:pipeline}
\end{figure}

Figure~\ref{fig:pipeline} shows the proposed framework for multi-view human image generation. The framework incorporates some design principles from Zero-1-to-3~\cite{zero1to3}, employing a view-conditioned diffusion model to synthesize novel views of an object. Unlike Zero-1-to-3, which processes one view at a time and may encounter view consistency issues, our framework processes all target views simultaneously and integrates features across views using the proposed mesh attention mechanism, detailed in Sec.~\ref{subsec:method.mesh_attention}. In addition, our framework incorporates the following designs to improve performance:
1) Keypoint conditioning, 
2) Resolution upscaling and choice of VAE, and 
3) Linear noise schedule.

\noindent \textbf{Keypoint Conditioning.} 
Our training dataset DNA-Rendering \cite{cheng2023dnarendering} comprises multiview-captured real human videos, offering a diverse range of poses. However, this also adds complexity to model learning.
To handle these complex poses, we propose incorporating detected skeleton keypoints of the target views into the model. 
Specifically, we add the keypoint features (after adjusting their spatial resolutions and channel numbers) to the U-Net features as a condition. 
By explicitly providing such keypoint conditioning, our model no longer needs to rely solely on camera parameters to estimate human poses in new views and can instead focus on ensuring cross-view consistency and generating detailed outputs.

\noindent \textbf{Resolution Upscaling and Choice of VAE.} As analyzed in Sec.~\ref{sec:intro} and Table \ref{tab:multiview_attn}, most multiview diffusion models are limited to a low-resolution of $256^2$, with only a few recent studies reaching $512^2$. As shown in Fig.~\ref{fig:vae_res} and \cref{tab:vae_res}, higher resolutions and improved VAE models are crucial for capturing highly detailed human data. To minimize cross-view inconsistencies and quality degradation caused by VAE reconstruction, we train our model using $1024\times1024$ images and use SDXL VAE~\cite{sdxl} in our framework. 

\noindent \textbf{Noise Schedule.} Following the recommendation from Zero123++~\cite{shi2023zero123++}, we use a linear schedule for the denoising process instead of a scaled-linear schedule to achieve better global consistency across multiple views.

\subsection{Inference} 

For in-the-wild image inputs, we crop the image according to the dataset setting, which we detail in \cref{sec:dataset} and \cref{suppsec:dataset}. We then apply ECON~\cite{xiu2023econ} to produce the clothed human mesh and the corresponding SMPL-X~\cite{smplx} parameters. 
Since ECON operates under an orthographic camera assumption, we first obtain a frontal perspective camera by optimization. Based on the ``Look-At" transformation, we assume the frontal camera has a fixed field of view (FoV) and directs at the pelvis. We optimize its position to align the rendered SMPL-X keypoints with those in the image. After that, we sample camera parameters that orbit around the human body, maintaining a fixed elevation and distance. 
We then project 3D keypoints to each view and perform rasterization and aggregation for mesh attention.

\begin{figure*}[t]
    \vspace{-0.5cm}
    \centering
    \includegraphics[width=0.95\textwidth]{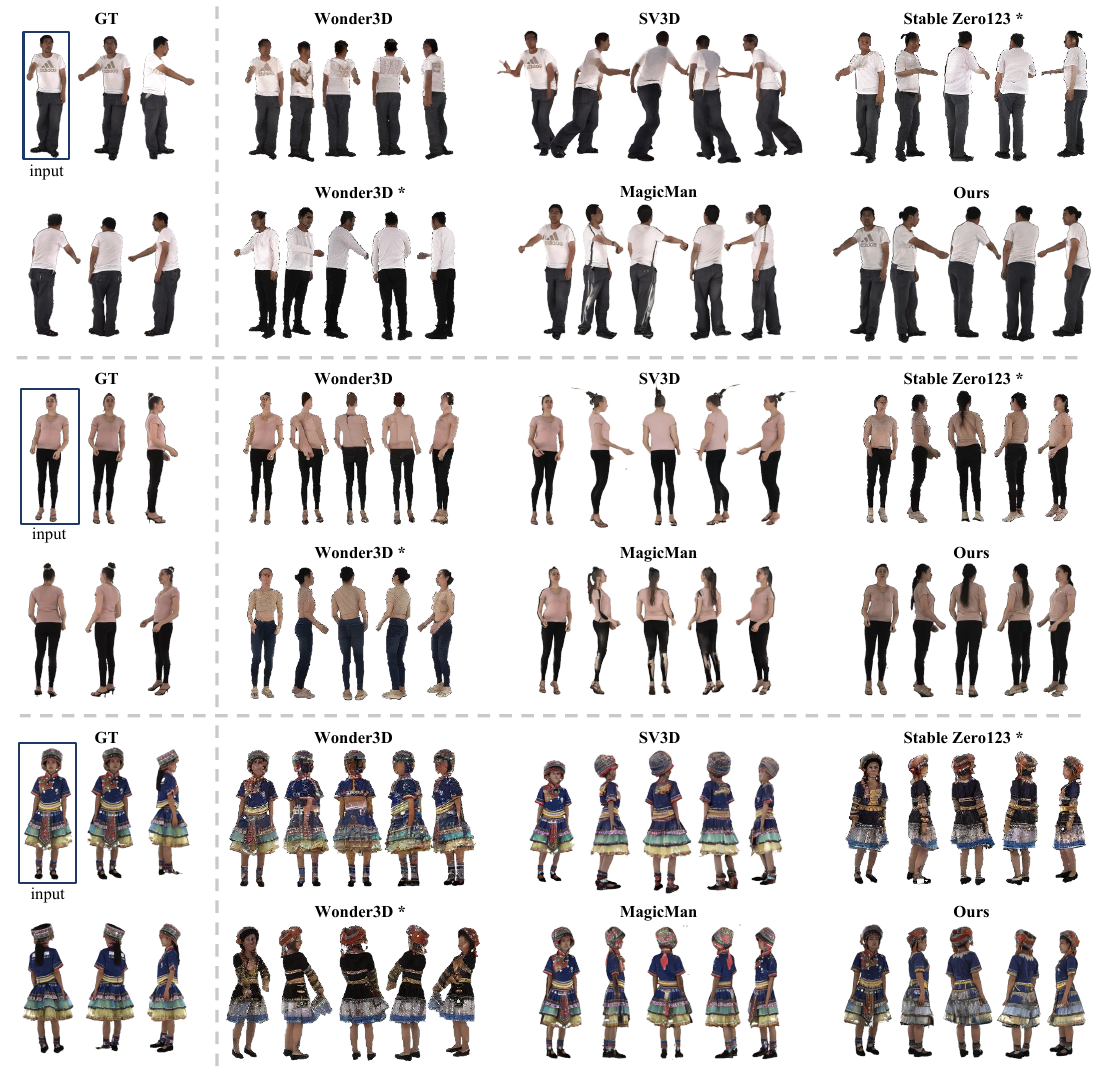}
    \caption{\textbf{Qualitative Results.} MEAT (Ours) demonstrates significant advantages in resolution, detail, and cross-view consistency in novel view synthesis tasks. * Methods are re-trained on the DNA-Rendering dataset for fair comparison. Please \textbf{zoom in} for details.}
    \vspace{-0.3cm}
    \label{fig:main_qualitative}
\end{figure*}
\section{Adapting DNA-Rendering for Training}
\label{sec:dataset}

Although the DNA-Rendering \cite{cheng2023dnarendering} dataset provides multiview images with resolutions exceeding $2000^2$ and an extensive range of human poses, the multiview setting brings additional challenges. Here we briefly describe how we deal with centric mesh adaptation, and image cropping with camera calibration. Details can be found in \cref{suppsec:dataset}.

\noindent \textbf{Mesh Adaptation.} To reduce the quality gap of the centric mesh between training and inference, we use the monocular-reconstructed mesh from a pre-selected frontal image for training. We choose PIFuHD\cite{saito2020pifuhd} for its balance of speed and quality. To align this mesh with the dataset's established calibration system, we need to compute a transformation $\text{TF}$ for the mesh, so that reference view pixels can reach matching points in adjacent views after projection and reprojection. We use RoMa~\cite{edstedt2024roma} to detect all feature-matching pairs and apply gradient descent to solve $\text{TF}$.

\noindent \textbf{Image Cropping.} To apply the simplified camera representation with azimuth and elevation, we need to simulate the results of capturing images from cameras arranged on an object-centric sphere. We use a camera series placed at the same altitude and assume the subject has the same height in each pixel plane. Then we align the pelvis joint from SMPL-X to the center of the pixel grid. We set the cropping radius to $1.3\times$ the maximum height difference between the keypoint and pelvis and resized the cropped images to the same resolution. Since only cropping and resizing are involved, we only need to adjust the principal point coordinates in the camera intrinsics and normalize the camera to the NDC (Normalized Device Coordinate) system. 

\section{Experiments}
\label{sec:experiments}

We compare our method with Stable Zero123 \cite{stable_zero123}, SyncDreamer \cite{liusyncdreamer}, Wonder3D \cite{long2024wonder3d}, SV3D \cite{voleti2025sv3d}, and MagicMan \cite{he2024magicman} on DNA-Rendering \cite{cheng2023dnarendering} Part 1. MagicMan is only compared qualitatively as its preset views cannot align with the test set. For quantitative comparison, we report benchmark results on both 256 and 1024 resolution, covering reconstruction metrics (PSNR, SSIM \cite{wang2004image_ssim}, and LPIPS \cite{zhang2018unreasonable_lpips}), generation quality metrics (FID and Patch-FID), and a cross-view consistency metric PPLC proposed by Free3D \cite{zheng2024free3d}. All the details are specified in \cref{suppsec:implementation}.

\subsection{Main Results}
\label{subsec:experiments.main_results}

\begingroup
\setlength{\tabcolsep}{5.5pt}
\begin{table*}[t]
\vspace{-0.5cm}
\caption{\textbf{Main Quantitative Results.} We highlight the best value in \colorbox{pearDark!20}{blue}, and the second-best value in \colorbox{mycolor_green}{green}. ``Infer." means we use the open-source checkpoint. For the retrained baselines, we also provide a version with keypoints conditioning for a fair comparison.}
\vspace{-10pt}
{\fontsize{8.9pt}{11pt}\selectfont
\begin{tabularx}{\textwidth}{llc|cccc|ccccc}
\toprule
\multirow{2}{*}{Method}   & \multirow{2}{*}{Type}  & \multirow{2}{*}{Res.}       & \multicolumn{4}{c|}{\textbf{1024}}  & \multicolumn{5}{c}{\textbf{256}} \\ 
                          &                        &                             & PSNR $\uparrow$ & SSIM $\uparrow$ & LPIPS $\downarrow$ & P-FID $\downarrow$ & PSNR $\uparrow$ & SSIM $\uparrow$ & LPIPS $\downarrow$ & FID $\downarrow$ & PPLC $\downarrow$                     \\ \midrule
Stable Zero123 \cite{stable_zero123}             & Infer.                 & 256                         & 9.039 & 0.7839 & 0.3299 & 74.24 & 9.056 & 0.7033 & 0.3966 & 55.16 & 0.4549          \\
SyncDreamer \cite{liusyncdreamer}                & Infer.                 & 256                         & 12.12 & 0.8653 & 0.2331 & 102.8 & 12.13 & 0.7998 & 0.3231 & 71.42 & 0.2017          \\
Wonder3D \cite{long2024wonder3d}                 & Infer.                 & 256                         & 16.58 & 0.9084 & 0.1456 & 59.79 & 16.68 & 0.8649 & 0.1359 & 39.32 & \tablesecond0.0897 \\ 
SV3D \cite{voleti2025sv3d}                       & Infer.                 & 578                         & 13.32 & 0.8843 & 0.1830 & \tablesecond24.99 & 13.43 & 0.8175 & 0.2372 & \tablesecond20.14 & 0.1333          \\ \midrule
Stable Zero123 \cite{stable_zero123}             & Train                  & 256                         & 17.52 & 0.9139 & 0.1345 & 62.71 & 17.62 & 0.8768 & 0.1173 & 34.53 & 0.1010           \\
\quad\texttt{+ kpts.} & Train & 256 & \tablesecond19.08 & 0.9201 & \tablesecond0.1234 & 63.39 & 19.22 & 0.8912 & 0.0941 & 33.64 & 0.0992 \\
Wonder3D \cite{long2024wonder3d}                 & Train                  & 256                         & 16.73 & 0.9081 & 0.1449 & 67.11 & 16.82 & 0.8684 & 0.1356 & 47.59 & 0.1042                      \\ 
\quad\texttt{+ kpts.} & Train & 256 & \tablefirst19.35 & \tablesecond0.9205 & 0.1239 & 72.40 & \tablefirst19.51 & \tablesecond0.8957 & \tablesecond0.0931 & 51.75 & \tablefirst0.0895 \\ \midrule
MEAT (Ours)                                      & 1-stage                & 1024                        & 18.91 & \tablefirst0.9271 & \tablefirst0.0751 & \tablefirst10.60 & \tablesecond19.41 & \tablefirst0.9043 & \tablefirst0.0791 & \tablefirst16.56 & 0.0991           \\ \bottomrule
\end{tabularx}
}
\label{tab:quant_main}
\vspace{-0.4cm}
\end{table*}
\endgroup

\noindent\textbf{Quantitative.} \Cref{tab:quant_main} presents the quantitative comparison with the baselines. For each method, we generate 16 pre-set viewpoints and compare them with the ground-truth images. We add keypoint conditions to the retrained Stable-Zero123 \cite{zero1to3, stable_zero123} and Wonder3D \cite{long2024wonder3d} with a similar scheme as described in \cref{subsec:method.human_multiview_diffusion} for a fair comparison. Our method achieves the best results across both resolutions in reconstruction metrics and leads in generation quality. Notably, MEAT significantly outperforms existing methods on the Patch-FID and LPIPS metrics, highlighting the value of megapixel-resolution training. For cross-view consistency metric (PPLC), Wonder3D achieves the best performance, with our method closely following. The results of Wonder3D highlight the significant improvement in cross-view consistency made possible by combining cross-domain attention. However, it is highly memory-intensive and difficult to scale to megapixel resolutions. In contrast, our method is much more efficient. We also find that keypoints conditioning significantly improves the numerical metrics of both baselines, but they still lag far behind our method in (P-)FID and LPIPS, due to low resolution.

\noindent\textbf{Qualitative.} \cref{fig:main_qualitative} shows the qualitative comparison with other baselines. All methods operating at $256^2$ resolution fail to produce facial details, and their texture clarity is noticeably inferior to that of MEAT. The pre-trained Wonder3D frequently generates duplicate back views with limited perspective variation, potentially giving it an unfair advantage in the PPLC metric. SV3D shows a clear improvement in resolution but falls short of our method in geometric consistency, lacking perceptual awareness of human structure. MagicMan, as a concurrent work, stands out among the baselines but still struggles with visible artifacts and incomplete limbs when generating side views (\eg, in the third example). Our method achieves high-resolution, detail-rich, and view-consistent human novel view synthesis. More examples are in \cref{suppsec:results}.

\subsection{Ablations and Discussions}
\label{subsec:experiments.ablation}

\begin{figure}[t]
    \centering
    \includegraphics[width=\columnwidth]{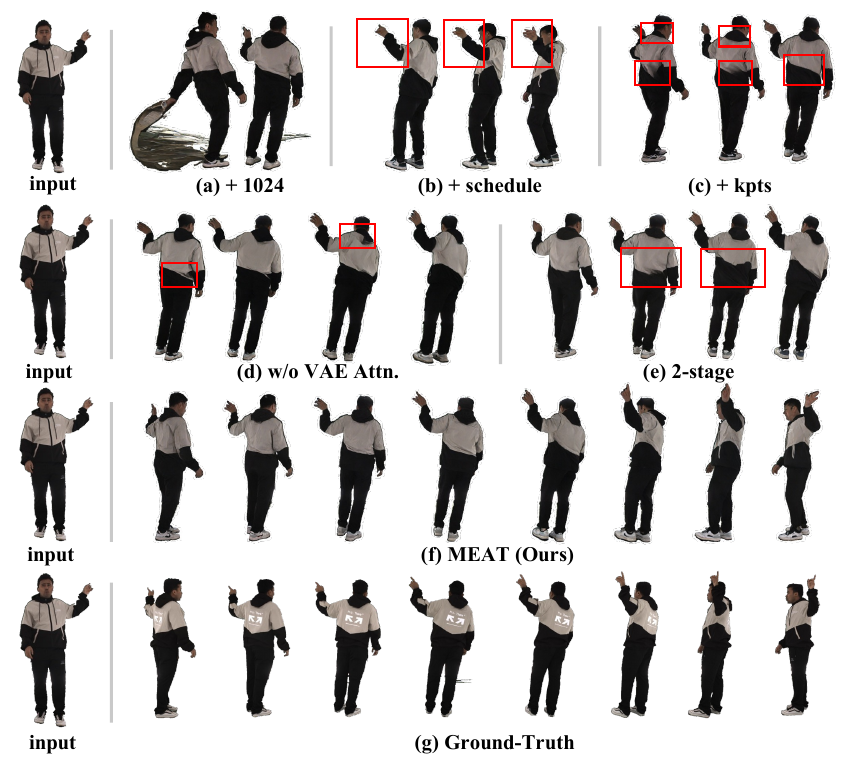}
    \caption{\textbf{Qualitative Ablation.} MEAT achieves the best cross-view consistency.}
    \label{fig:ablation_qualitative}
    \vspace{-0.6cm}
\end{figure}
\begingroup
\setlength{\tabcolsep}{3pt}
\begin{table}[t]
\caption{\textbf{Quantitative Ablation.} Best value in \colorbox{pearDark!20}{blue}, second-best in \colorbox{mycolor_green}{green}. $^{*}$Here PPLC is calculated on $1024\times1024$ resolution.}
\vspace{-10pt}
{\fontsize{8.0pt}{11pt}\selectfont
\begin{tabularx}{\columnwidth}{l|cccccc}
\toprule
Method                 & PSNR $\uparrow$ & SSIM $\uparrow$ & LPIPS $\downarrow$ & FID $\downarrow$ & P-FID $\downarrow$     & $\text{PPLC}^{*}\downarrow$ \\ \midrule
SZ123 - 256            & 17.51           & 0.9139          & 0.1344             & 23.71            & 62.71                  & $-$            \\
+ res. 1024            & 14.41           & 0.8873          & 0.1480             & 21.41            & 14.56                  & 0.1805            \\
+ schedule             & 16.56           & 0.9114          & 0.1023             & 16.81            & 11.21                  & 0.1170            \\
+ keypoints            & 18.78           & 0.9238          & 0.0776             & \tablesecond16.19            & 10.79                  & 0.0995            \\
$\times$ VAE Attn      & 18.50           & 0.9233          & 0.0788             & 16.91            & 10.76                  & 0.0981            \\ 
2-stage                & \tablefirst19.11  & \tablesecond0.9266          & \tablesecond0.0755             & \tablefirst15.37   & \tablefirst9.983        & \tablesecond0.0973   \\ 
\textbf{Ours}          & \tablesecond18.91           & \tablefirst0.9271 & \tablefirst0.0751    & 17.08            & \tablesecond10.60                  & \tablefirst0.0928   \\ \bottomrule
\end{tabularx}
}
\label{tab:quant_ablation}
\vspace{-0.4cm}
\end{table}
\endgroup

The qualitative and quantitative ablation results are shown in \cref{fig:ablation_qualitative} and \Cref{tab:quant_ablation}, respectively.

\noindent\textbf{Resolution Upscaling.} Directly increasing the training resolution to 1024 causes Stable Zero123 to generate numerous artifacts, as shown in \cref{fig:ablation_qualitative}-(a). Adjusting the noise scheduler to reduce the SNR at the beginning of the denoising process is key to mitigating this issue (see \cref{fig:ablation_qualitative}-(b)).

\noindent\textbf{Keypoint Conditioning.} Without the keypoint condition, in \cref{fig:ablation_qualitative}-(b), the generated results show noticeable misalignment in the left arm, when compared against the reference view and ground truth. The keypoint conditioning reduces the model's difficulty in understanding the human geometric structure. 

\noindent\textbf{Mesh Attention.} Adding only keypoint conditioning does not ensure cross-view consistent texture generation, as each view is still generated independently (see \cref{fig:ablation_qualitative}-(c)). Mesh attention is the key to address the consistency issue. We compared three variants with mesh attention. Models without VAE attention tend to produce local consistency anomalies, as is shown in \cref{fig:ablation_qualitative}-(d). We examine a 2-stage training strategy for MEAT, where we first train the U-Net without mesh attention for 100k iterations, then freeze these parameters and train the mesh attention blocks for another 50k iterations. This model shows slightly better generation quality in terms of FID, but exhibits noticeable cross-view inconsistency, especially in texture patterns such as color blocks. See \cref{fig:ablation_qualitative}-(e). As is reflected by \cref{fig:ablation_qualitative}-(f) and the PPLC metric, 1-stage-trained MEAT shows the best cross-view consistency. Check \cref{suppfig:crossview} for more comparison.

\subsection{More Analysis}
\label{subsec:experiments.analysis}

\begingroup
\setlength{\tabcolsep}{2.2pt}
\begin{table}[t!]
\vspace{-0.5cm}
\caption{\textbf{Other Attention Schemes.} Best value in \colorbox{pearDark!20}{blue}. Memory reported in \textit{GB} as Train/Inference. {\color{red}red} for estimation due to OOM. All metrics on $1024^2$ resolution.}
\vspace{-10pt}
{\fontsize{8pt}{11pt}\selectfont
\begin{tabularx}{\columnwidth}{l|cc|ccccc}
\toprule
Attn. & {\scriptsize $\text{Mem}_{512}$} & {\scriptsize $\text{Mem}_{1024}$} & {\scriptsize PSNR $\uparrow$} & {\scriptsize SSIM $\uparrow$} & {\scriptsize LPIPS $\downarrow$} & {\scriptsize P-FID $\downarrow$} & {\scriptsize PPLC $\downarrow$} \\ \midrule
Dense         & 41/41 & {\color{red}186/471} & 18.77 & 0.9226 & 0.1004 & 37.34 & 0.1049 \\
Epipolar      & 51/46 & {\color{red}154/294} & 18.55 & 0.9217 & 0.1017 & 38.37 & 0.1063 \\
\textbf{Mesh} & \tablefirst{37/24} & \tablefirst{68/52} & \tablefirst18.91 & \tablefirst0.9271 & \tablefirst0.0751 & \tablefirst10.60 & \tablefirst0.0929 \\ \bottomrule
\end{tabularx}
}
\label{tab:quant_attn}
\vspace{-0.2cm}
\end{table}
\endgroup

\noindent\textbf{Other Attention Schemes.} To further demonstrate the importance of mesh attention for high-resolution multiview generation, we replaced the mesh attention module in MEAT with alternative attention schemes summarized in \cref{tab:multiview_attn}, except row-wise attention, which is designed for orthographic cameras of equal heights, incompatible with our dataset. \cref{tab:quant_attn} shows the memory consumption and quantitative results. Mesh attention is much more memory-efficient in both training and inference, and is the only one able to train at 1024 with leading quantitative results.

\noindent\textbf{Necessity of 1024.} Main paper \cref{fig:vae_res} shows that VAE can only achieve satisfactory reconstruction on 1024. The quantitative comparison of VAE compression at different resolutions is presented in \cref{tab:vae_res}.
$\text{ID}_{GT}$ is ArcFace \cite{deng2019arcface} embedding similarity between the VAE output and the ground-truth image. $\text{ID}_{CV}$ is the cross-view similarity of the VAE results.
VAE performs poorly at low resolution in preserving facial features and exhibits greater cross-view inconsistency.

\begingroup
\setlength{\tabcolsep}{3pt}
\begin{table}[t]
\caption{\textbf{VAE Compression Quality.} Best value in \colorbox{pearDark!20}{blue}. All metrics are calculated on $1024\times1024$ resolution.}
\vspace{-10pt}
{\fontsize{9pt}{11pt}\selectfont
\begin{tabularx}{\columnwidth}{l|cccc|cc}
\toprule
Res. & PSNR $\uparrow$ & SSIM $\uparrow$ & LPIPS $\downarrow$ & P-FID $\downarrow$ & $\text{ID}_{GT}\uparrow$ & $\text{ID}_{CV}\uparrow$ \\ \midrule
256  & 28.02 & 0.9510 & 0.0914 & 53.14 & 0.2019 & 0.3917 \\
512  & 30.99 & 0.9689 & 0.0421 & 14.98 & 0.3984 & 0.3819 \\
1024 & \tablefirst35.59 & \tablefirst0.9866 & \tablefirst0.0063 & \tablefirst1.584 & \tablefirst0.7552 & \tablefirst0.5041 \\ \bottomrule
\end{tabularx}
}
\label{tab:vae_res}
\vspace{-0.5cm}
\end{table}
\endgroup

\noindent\textbf{Inpact of Keypoint Accuracy.} In the MEAT pipeline, keypoints conditions are mapped to 2D from 3D coordinates and then plotted. The 3D coordinates are computed from the SMPL-X \cite{smpl, smplx} parameters. For quantitative experiments, we use the ground-truth SMPL-X. For inference, we use PIXIE \cite{feng2021pixie} for prediction and ECON \cite{xiu2023econ} for optimization. The inference approach for obtaining SMPL-X inevitably introduces errors. Here we analyze the potential impact of these errors on the generation quality.

To simulate the impact of erroneous keypoints, we add Gaussian noise to ground-truth SMPL-X parameters, resulting in $60.79mm$ MPJPE. More specifically, we keep the main node (pelvis) unchanged and divide the other joints into two classes, main-body and hands. For each class, since each node is parametrized as an axis angle representation, we add Gaussian noise to the axis and the angle independently, with standard variance $\sigma$ and $\pi\sigma$ respectively. The main-body class has $\sigma_{main}=0.06$, while the hand class has $\sigma_{hand}=0.2$. This is to simulate the common SMPL-X prediction error distribution, where hands are usually more twisted. As is shown in \cref{tab:error_smplx}, the inference result is slightly declined due to inaccurate conditions, but still significantly outperforms the baselines.

\begingroup
\setlength{\tabcolsep}{3pt}
\begin{table}[t]
\vspace{-0.5cm}
\caption{\textbf{Impact of Erroneous Keypoints.} Best value in \colorbox{pearDark!20}{blue}. Erroneous keypoints degrade the numerical performance of our generated results, but do not compromise the overall superiority.}
\vspace{-10pt}
{\fontsize{8.0pt}{11pt}\selectfont
\begin{tabularx}{\columnwidth}{l|cccccc}
\toprule
Keypoints         & PSNR $\uparrow$ & SSIM $\uparrow$ & LPIPS $\downarrow$ & FID $\downarrow$ & P-FID $\downarrow$     & $\text{PPLC}\downarrow$ \\ \midrule
Perturbed  & 17.64 & 0.9201 & 0.0902 & 18.87 & 11.17 & 0.0992            \\
Ground-Truth    & \tablefirst18.91 & \tablefirst0.9271 & \tablefirst0.0751 & \tablefirst17.08 & \tablefirst10.60 & \tablefirst0.0928 \\ \bottomrule
\end{tabularx}
}
\label{tab:error_smplx}
\vspace{-0.2cm}
\end{table}
\endgroup

\noindent\textbf{Generalization to Poses.} The DNA-Rendering dataset categorizes each multiview video with labels indicating the challenging property associated with it. A subset of the data is tagged as "motion" and is further subdivided into \textit{simple}, \textit{medium}, and \textit{hard}. Here we show categorized results of our method and baselines on data tagged as ``motion", as a proxy of pose difficulty. \cref{tab:generalization_pose} shows that, more challenging poses uniformly degrade LPIPS and PPLC of all methods, but do not affect our leadership.

\begingroup
\setlength{\tabcolsep}{2.5pt}
\begin{table}[t]
\caption{\textbf{Generalization to Poses.} Best value in \colorbox{pearDark!20}{blue}. All metrics on $1024^2$ resolution.}
\vspace{-10pt}
{\fontsize{7.9pt}{11pt}\selectfont
\begin{tabularx}{\columnwidth}{l|ccc|ccc|ccc}
\toprule
\multirow{2}{*}{Method} & \multicolumn{3}{c|}{\textbf{motion simple}} & \multicolumn{3}{c|}{\textbf{motion medium}} & \multicolumn{3}{c}{\textbf{motion hard}} \\
 & {\scriptsize LPIPS} & {\scriptsize P-FID} & {\scriptsize PPLC} & {\scriptsize LPIPS} & {\scriptsize P-FID} & {\scriptsize PPLC} & {\scriptsize LPIPS} & {\scriptsize P-FID} & {\scriptsize PPLC} \\ \midrule
SZ123         & 0.118 & 81.78 & 0.090 & 0.126 & 77.61 & 0.098 & 0.148 & 69.19 & 0.122 \\
WD3D          & 0.124 & 90.27 & 0.089 & 0.133 & 88.36 & 0.097 & 0.159 & 74.29 & 0.122 \\
\textbf{MEAT} & \tablefirst0.054 & \tablefirst19.40 & \tablefirst0.076 & \tablefirst0.061 & \tablefirst17.92 & \tablefirst0.082 & \tablefirst0.086 & \tablefirst16.41 & \tablefirst0.107 \\ \bottomrule
\end{tabularx}
}
\label{tab:generalization_pose}
\vspace{-0.5cm}
\end{table}
\endgroup

\section{Conclusion}
\label{sec:conclusion}

In this paper, we propose \textbf{MEAT}, a human-specific multiview diffusion model that generates dense novel views of humans on megapixels conditioned on a frontal image. 
Our proposed \textbf{mesh attention} uses the monocular-reconstructed human mesh as a coarse central geometric representation, establishing cross-view coordinate correspondences through rasterization and projection. It enables highly memory-efficient cross-view attention, which overcomes the high complexity that hinders increasing the resolution to $1024^2$ for existing multiview attention methods. 
Through a series of techniques, we have, for the first time, enabled training a multiview diffusion model using multiview human motion videos, effectively enhancing the pose diversity of the training dataset.
Extensive experiments demonstrate that our generated multi-view human images exhibit significant advantages in cross-view consistency, clarity, and detail quality.

\small \noindent\textbf{Acknowledgement.} This work is supported by the National Research Foundation, Singapore under its AI Singapore Programme (AISG Award No: AISG2-PhD-2022-01-030), the RIE2020 Industry Alignment Fund Industry Collaboration Projects (IAF-ICP) Funding Initiative, as well as cash and in-kind contribution from the industry partner(s).

{
    \small
    \bibliographystyle{ieeenat_fullname}
    \bibliography{ref}
}

\clearpage
\section*{Appendix}
\label{sec:appendix}
In the supplementary material, we discuss further details and provide more results that are not included in the main paper. 
In \cref{suppsec:implementation}, we provide more details of our model setting and structure. 
In \cref{suppsec:dataset}, we discuss further details and provide visualization for our dataset processing pipeline. 
In \cref{suppsec:results}, we present more results on qualitative comparison with monocular reconstruction methods and illustration of our cross-view consistency preservation ability.

\appendix
\section{Implementation Details}
\label{suppsec:implementation}

In this section, we specify the details regarding the model implementation and the experiment settings.

\subsection{Model Implementation}

\noindent\textbf{Keypoints Conditioning.} We use a small 3-layer convolutional network to process the keypoints condition, downsampling the keypoints visualization image by 8x and aligning it with the channel of the denoiser U-Net after the \texttt{conv\_in} block. Each downsampling is achieved with two convolutional layers. The final output is processed with a \texttt{conv\_out} convolutional layer, which is zero-initialized to allow this condition to be smoothly integrated into the U-Net. We found that an additional branch like ControlNet-\cite{zhang2023controlnet} is unnecessary. Directly adding the processed condition to the U-Net features yields satisfactory training results.

\noindent\textbf{VAE Feature Encoder.} The VAE feature encoder is very similar to the diffusion U-Net down-sampling blocks without \textit{Attention} layers. At each resolution scale, there are 2 layers of \texttt{ResnetDownsampleBlock2D}, whose number of channels is matched with that in the U-Net. We use the last features before down-sampling in each residual block to be fused into the U-Net through VAE attention. 

\noindent\textbf{Implementation Details.} Our model is initialized with Stable Zero123 \cite{stable_zero123} pretrained weights, and optimized using $\epsilon\text{-prediction}$. Notably, since the SDXL-VAE~\cite{sdxl} can produce \texttt{NaN} under \texttt{fp16} precision, we utilize the \texttt{fp16-fix} version~\cite{sdxl-vae-fp16-fix} to support mixed-precision training. Our model supports sparse-view training. We randomly sample seven views, including the reference, in each training batch. The batch size on each GPU is 1, and we use 8 NVIDIA-A100-80GB GPUs to train 150,000 iterations without gradient accumulation, which takes about 7 days. Our model can generate 16 views simultaneously during inference. It employs a \texttt{Trailing} sample steps selection method to minimize the signal-to-noise ratio (SNR) at the beginning of the denoising process. We use DDIM sampler with 50 steps and a CFG scale of 3.0.

\subsection{Experiment Setting}

\noindent\textbf{Baselines.} For quantitative experiments, we compare our method with Stable Zero123 \cite{stable_zero123}, SyncDreamer \cite{liusyncdreamer}, Wonder3D \cite{long2024wonder3d}, and SV3D \cite{voleti2025sv3d}. For Wonder3D with pretrained weights, as it generates six views at a time, we split the 15 non-reference test views into three batches, each combined with the reference view for the generation. We re-train Stable Zero123 and Wonder3D on DNA-Rendering at the resolution of $256\times256$. Wonder3D is only trained in the color domain since ground-truth normal maps are not available. We only compare the results of MagicMan \cite{he2024magicman} qualitatively as its preset views cannot align with the test setting.

\noindent\textbf{Metrics.}
Since most of the previous multi-view diffusion models only generate at a resolution of 256, we also resize our results to calculate metrics at this resolution for fair comparison. Moreover, to show the advantage of high-resolution generation, we also compute metrics at a resolution of 1024.
For both resolutions, we include PSNR, SSIM \cite{wang2004image_ssim}, and LPIPS \cite{zhang2018unreasonable_lpips} metrics to compare the generated results with the ground-truth images. For the 1024 category, we use Patch-FID (P-FID) \cite{chai2022anyresgan, fu2023unitedhuman, li2024cosmicman} instead of FID \cite{FID} as a metric for generation quality. FID resizes images to $299\times299$ before calculation, which does not reflect MEAT's advantage at high resolutions. Instead, we split each image into a $4\times4$ grid of $256\times256$ patches and select the middle two columns, yielding eight patches per image. The calculation is based on the patch set. In the 256 categories, we also use the PPLC metric proposed by Free3D \cite{zheng2024free3d} to evaluate cross-view consistency in multiview generation. We exclude it in the 1024 category because upsized blurry results gain an unfair advantage in this metric.
\section{DNA-Rendering for Multiview Generation}
\label{suppsec:dataset}

In this section, we present the full details of the novel ideas proposed to harness multiview human video dataset DNA-Rendering~\cite{cheng2023dnarendering} for multiview diffusion training. We construct our training data using the multiview human dataset DNA-Rendering \cite{cheng2023dnarendering}, which provides 15 FPS multiview videos of human motion. By sampling one set of frames every five frames, we generate over 20,000 sets of multiview images. The first partition, containing 2,000 samples, is reserved for testing, while the second partition is used for training. 
While this larger dataset offers a significant advantage, the multiview setting brings additional challenges. We address three primary issues: (1) selecting the front view for monocular reconstruction, (2) adapting the monocular reconstructed mesh to the calibrated coordinate system, and (3) cropping the images with corresponding adjustments to the camera calibration parameters.

\subsection{Frontal Camera Selection}

For each frame of multiview images in the DNA-Rendering~\cite{cheng2023dnarendering} dataset, we need to first determine which view is the ``frontal" one. This config is utilized in monocular reconstruction, training views sampling, and inference. Since the dataset provides the SMPL-X coefficients and camera calibration parameters $R_v$ and $T_v$ for each view, we can derive the global orientation $\bm{d}$ of the human body, the 3D coordinates $\bm{G}$ of the pelvis, and the camera coordinates $\bm{C}_{v}$, where 
\begin{equation*}
    \label{eq:camera_location}
    \bm{C}_{v}=-R_{v}^{-1}T_{v}.
\end{equation*}
We define the frontal view as the viewpoint where the angle between the line connecting the camera's optical center to the pelvis and the global orientation is minimized, \ie
\begin{equation}
    \label{eq:frontal}
    \text{front view} \leftarrow \operatorname*{arg\,max}_{v} \frac{\bm{d}\cdot\bm{G}\bm{C}_{v}}{\|\bm{d}\| \|\bm{G}\bm{C}_{v}\|}
\end{equation}

\subsection{Mesh Adaptation}

\begin{figure*}[h!]
    \centering
    \begin{subfigure}[b]{\textwidth}
        \centering
        \includegraphics[width=0.9\textwidth]{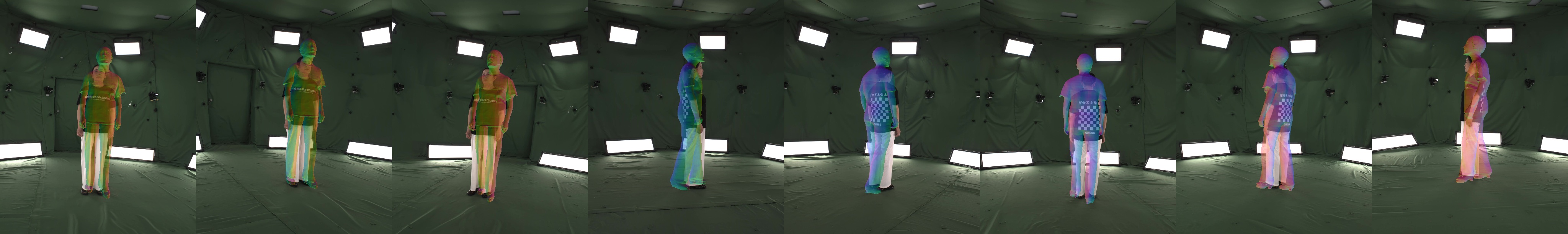}
        \caption{Mesh location before adaptation.}
    \end{subfigure}
    
    \begin{subfigure}[b]{\textwidth}
        \centering
        \includegraphics[width=0.9\textwidth]{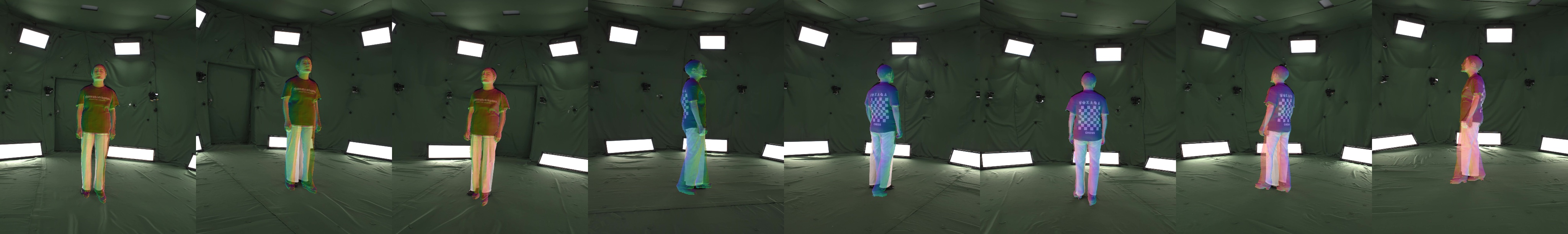}
        \caption{Mesh location after adaptation.}
    \end{subfigure}
    
    \caption{\textbf{Mesh Adaptation.} Although the monocular reconstructed human mesh inevitably exhibits certain deviations from the ground truth, our mesh adaptation method can robustly align it to the dataset's coordinate system. Our MEAT model, trained using this data, effectively mitigates the interference of geometric noise in human meshes during multi-view image generation.}
    \label{suppfig:mesh_adaptation} 
\end{figure*}

To ensure consistent mesh quality during both training and inference and to prevent the model from overly relying on the accuracy of the centric geometric representation, we use monocular reconstruction from the selected frontal image above to extract the centric mesh for training. We use PIFuHD\cite{saito2020pifuhd} for its balance of speed and quality. However, monocular reconstruction typically assumes a specific position and orthographic projection for the frontal camera, which differs from our dataset where the frontal camera is perspective and can be positioned variably. Consequently, we need to determine a transformation $\text{TF}$ to align the mesh with the world coordinate system of the dataset. 

Our adaptation approach is based on the following rule: $\bm{P_p}$ of each pixel $\bm{p}$ in the reference view, after transformation $\text{TF}$ and reprojection, should return to its original position in its own view and reach the feature-matching point in adjacent views. These two relationships establish an optimization objective for $\text{TF}$ with a unique optimal solution. We use RoMa~\cite{edstedt2024roma} to detect all feature-matching pairs and apply gradient descent to solve $\text{TF}$.

Specifically, we assume that the transformation $\text{TF}$ for each vertex $\bm{P}$ consists of a scaling $S$, rotation $R$, and translation $\bm{t}$:
\begin{align}
    \label{eq:vertex_transformation}
    S&=\mathrm{diag}(\bm{s}), \bm{s}=[s_x,s_y,s_z], \\
    R&=\texttt{rot6d}(\bm{c}_1, \bm{c}_2), \\
    p'&=\text{TF}(\bm{P})=R (S \bm{P}) + \bm{t}.
\end{align}
We use \texttt{rot6d} rotation representation \cite{zhou2019continuity_rot6d} for more stable optimization. We can then define the re-projection process $\tilde{\Pi}_v$ of a frontal-view pixel $\bm{p}$ into the view $v$.
\begin{equation}
    \label{eq:reproj}
    \tilde{\Pi}_{v}(\bm{p})=\Pi_{v}(\text{TF}(\bm{p}\rightarrow \bm{P})).
\end{equation}
Here $\bm{p}\rightarrow \bm{P}$ indicates the inverse orthographic rasterization process and $\Pi_v$ is the projection to view $v$ as is described in Eq.(6) in the main paper. Let $v=1$ be the frontal view. We use two types of alignment to build the optimization target: 
\begin{enumerate}
    \item $\tilde{\Pi}_{1}(\bm{p})$ - Pixels return to their original positions.
    \item $\tilde{\Pi}_{v}(\bm{p})$ - Pixel $\bm{p}$ on the frontal view is matched with pixel $\bm{q}_v$ on view $v$. 
\end{enumerate}
We use RoMa~\cite{edstedt2024roma} to detect such $(\bm{p}, \bm{q}_v)$ pairs. All the pixels $\bm{p}$ that do not intersect with the mesh are filtered out. The pixel values are normalized to $[0,1]$ based on the resolution of the raw image. Finally, we can solve the transformation $\text{TF}$ through:
\begin{equation}
    \label{eq:mesh_adaptation_target}
    \operatorname*{arg\,min}_{\bm{s},\bm{c}_1,\bm{c}_2,\bm{t}} \quad \sum_{\bm{p}} \| \bm{p} - \tilde{\Pi}_{1}(\bm{p}) \|_2^2 + \sum_{\bm{p},\bm{q}_v} \| \bm{q}_{v} - \tilde{\Pi}_{v}(\bm{p}) \|_2^2.
\end{equation}

We initialize these parameters with the assumption of zero translation, identical scaling, and an aligned coordinate system. It yields $\bm{s}_0=[1,1,1]$, $\bm{t}_0=\bm{0}$, and
\begin{equation}
    \label{eq:intialization}
    R_0 = \left(
    \begin{bmatrix}
    1 &  0 & 0 \\
    0 & -1 & 0 \\
    0 &  0 & -1
    \end{bmatrix}
    \cdot R_{v\text{=}1} \right)^{-1}
\end{equation}
Here $R_{v\text{=}1}$ is the calibrated extrinsic rotation matrix of the frontal camera in the DNA-Rendering~\cite{cheng2023dnarendering} dataset. DNA-Rendering adopts the \texttt{opencv} camera coordinate system convention, which has an opposite direction of $y\text{-axis}$ and $z\text{-axis}$. We show visualization results in \cref{suppfig:mesh_adaptation}.

\subsection{Image Cropping}

Existing multiview diffusion models place the object at the origin of the world coordinate system when rendering datasets, and position the camera on a fixed-radius sphere centered at this origin. This approach simplifies the viewpoint representation to just azimuth and elevation, reducing training complexity. 

During training, we use the 1-meter-high circular camera array of DNA-Rendering to simulate the zero-elevation rendered data. These cameras are all oriented toward the calibrated center of the world coordinate system. However, this center often does not align precisely with the person’s position, resulting in variable positioning within the images. This variability introduces ambiguity when using the camera representation of existing multiview diffusion models.

To address this issue, we propose cropping the images based on the pelvis position. We align the pelvis joint from SMPL-X in each frame to the center of the pixel grid. To maintain consistency with the spherical camera arrangement, we assume the subject has the same height in each pixel plane since all cameras have the same height. We set the cropping radius to $1.3\times$ the maximum height difference between any keypoint and the pelvis in each pixel plane:
\begin{equation}
    \label{eq:recropping}
    R_{v} = 1.3 \cdot \max_{\bm{P}} |\Pi_{v}(\bm{P})_{y} - \Pi_{v}(\bm{P}_{pelvis})_y|.
\end{equation}
The cropped images from each view are then resized to the same resolution. Since only cropping and resizing are involved, we only need to adjust the principal point coordinates in the camera intrinsics and normalize the camera to the NDC (Normalized Device Coordinate) system.

\section{More Results}
\label{suppsec:results}

\subsection{Cross-view Consistency Preservation}

\begin{figure*}[t]
    \centering
    \includegraphics[width=0.9\textwidth]{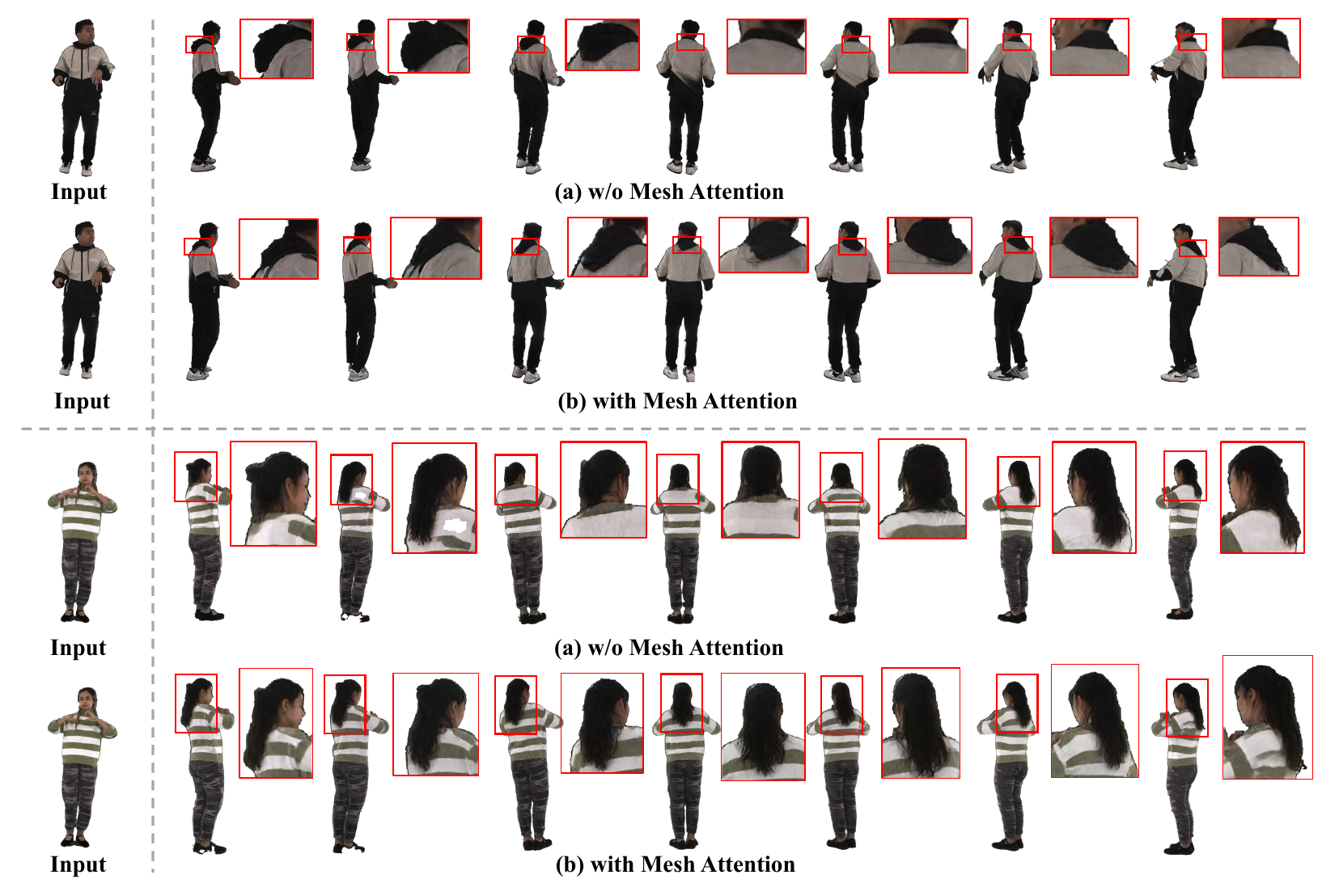}
    \caption{\textbf{Cross-view Consistency Preservation.} Models without mesh attention adhere to a one-view-at-a-time approach. Due to the stochastic nature of diffusion models, generating the backside often fails to maintain local structural consistency across different viewpoints. The mesh attention module significantly enhances the cross-view consistency preservation.}
    \label{suppfig:crossview}
\end{figure*}

We show the generated results of models with and without mesh attention modules in \cref{suppfig:crossview}. In the multiview diffusion model, the generation of front-facing regions leverages information from reference viewpoints, resulting in reduced randomness. Conversely, the generation of the backside relies more heavily on the model's generative capabilities, thereby exhibiting greater randomness inherent to diffusion models. As is shown in \cref{suppfig:crossview}, one-view-at-a-time models lacking mesh attention frequently make random selections among different modes in local structures, resulting in inconsistencies across viewpoints. The mesh attention module effectively mitigates this issue, achieving better cross-view consistency preservation.

\subsection{Monocular Reconstruction Methods}

\begin{figure*}[t]
    \centering
    \includegraphics[width=0.6\textwidth]{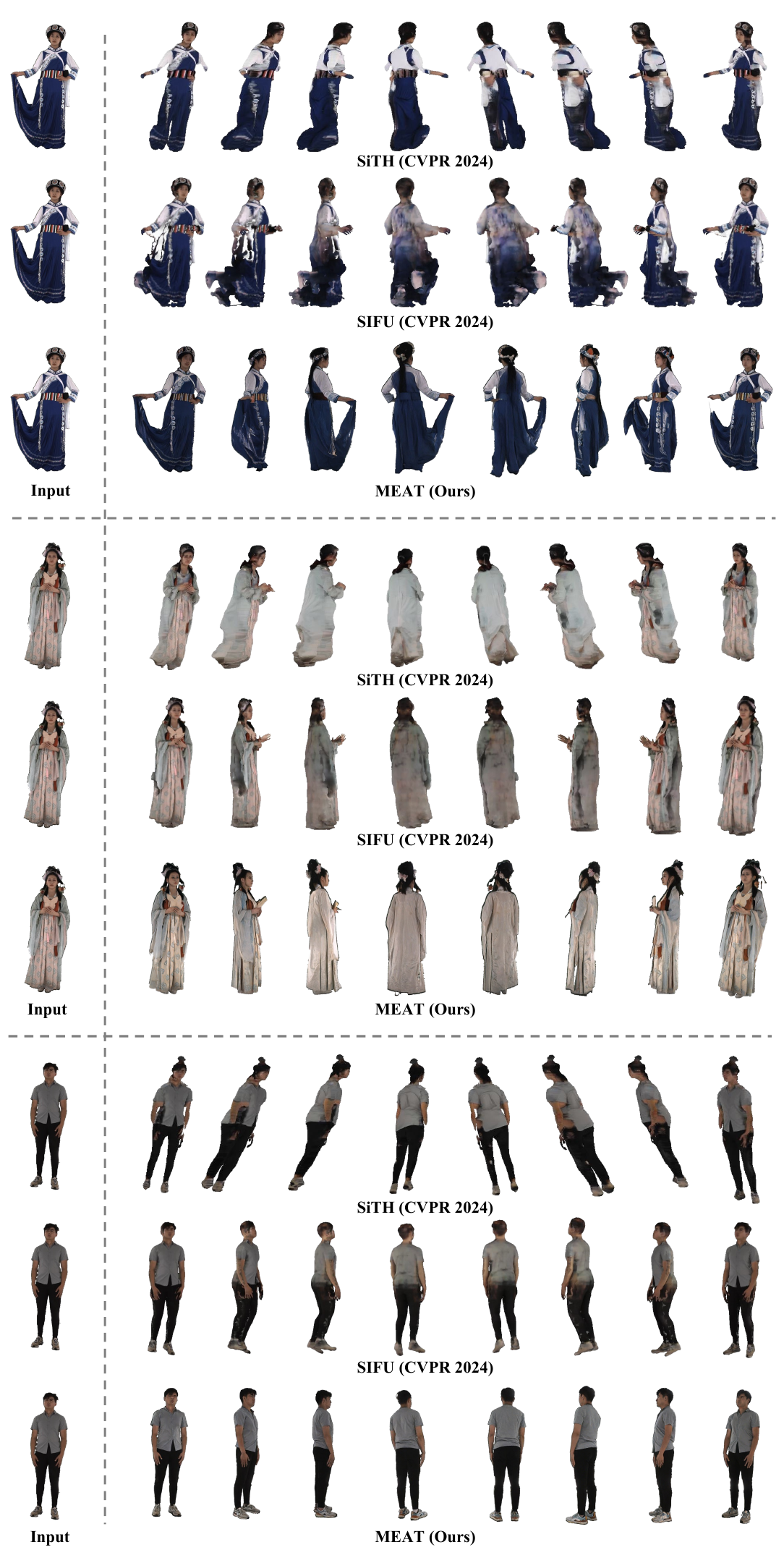}
    \caption{\textbf{Comparison with Monocular Reconstruction Methods.} In the novel view generation results for human bodies, compared to monocular reconstructed meshes, the multiview images generated by our MEAT diffusion model exhibit significant advantages in geometric plausibility, geometric details, texture details, and clarity. Please \textbf{zoom in} for details.}
    \label{suppfig:monocular}
\end{figure*}

In this section, we compare the novel view generation results of our MEAT diffusion model with monocular reconstruction methods like SiTH~\cite{ho2024sith} and SIFU~\cite{zhang2024sifu}. The qualitative comparison results are shown in \cref{suppfig:monocular}. For monocular reconstruction methods, novel view images are rendered from textured human meshes, thereby inherently ensuring perfect cross-view consistency. 

However, due to the challenges associated with accurate geometric estimation, monocular reconstructed human meshes often exhibit reduced realism when dealing with relatively loose clothing, thus the results after texture mapping are unsatisfactory. Our MEAT model utilizes such coarse human meshes solely as a medium for cross-view feature fusion; the generated images themselves are not rendered from any explicit geometric representations, resulting in a noticeable enhancement in realism.

\end{document}